\documentclass[acmtog]{acmart}
\acmSubmissionID{344}

\usepackage{booktabs} 
\usepackage{multirow}
\usepackage{subfigure}
\usepackage{graphicx}
\usepackage{overpic}
\usepackage{bbm}

\usepackage[ruled]{algorithm2e} 

\SetAlFnt{\small}
\SetAlCapFnt{\small}
\SetAlCapNameFnt{\small}
\SetAlCapHSkip{0pt}

\usepackage{color}
\definecolor{blue}{rgb}{0,0,1}
\definecolor{red}{rgb}{1,0,0}
\definecolor{green}{rgb}{0,.5,0}
\definecolor{orange}{rgb}{0.75, 0.4, 0}
\definecolor{teal}{rgb}{0.0, 0.4, 0.4}
\definecolor{purple}{rgb}{0.65,0,0.65}
\definecolor{gray}{rgb}{0.3,0.3,0.3}
\definecolor{black}{rgb}{1,1,1}

\newcommand{\rh}[1]{#1}
\newcommand{\rev}[1]{#1}
\newcommand{\ok}[1]{#1}

\acmJournal{TOG}

\setcopyright{acmcopyright}\acmJournal{TOG}
\acmYear{2022}\acmVolume{41}\acmNumber{6}\acmArticle{198}\acmMonth{12} \acmDOI{10.1145/3550454.3555483}

\begin{document}
\title{Asynchronous Collaborative Autoscanning with Mode Switching for Multi-Robot Scene Reconstruction}

\author{Junfu Guo}
\orcid{0000-0002-2217-5069}
\affiliation{%
   \institution{University of Science and Technology of China}
   \country{China}
}

\author{Changhao Li}
\orcid{0000-0003-0850-8987}
\affiliation{%
   \institution{University of Science and Technology of China}
   \country{China}
}

\author{Xi Xia}
\orcid{0000-0002-3396-9243}
\affiliation{%
   \institution{University of Science and Technology of China}
   \country{China}
}

\author{Ruizhen Hu}
\authornote{Corresponding author: Ruizhen Hu (ruizhen.hu@gmail.com)}
\orcid{0000-0002-6798-0336}
\affiliation{%
   \institution{Shenzhen University}
   \country{China}
}

\author{Ligang Liu}
\orcid{0000-0003-4352-1431}
\affiliation{%
   \institution{University of Science and Technology of China}
   \country{China}
}

\begin{teaserfigure}
    \begin{overpic}[width=1.0\textwidth,tics=10]{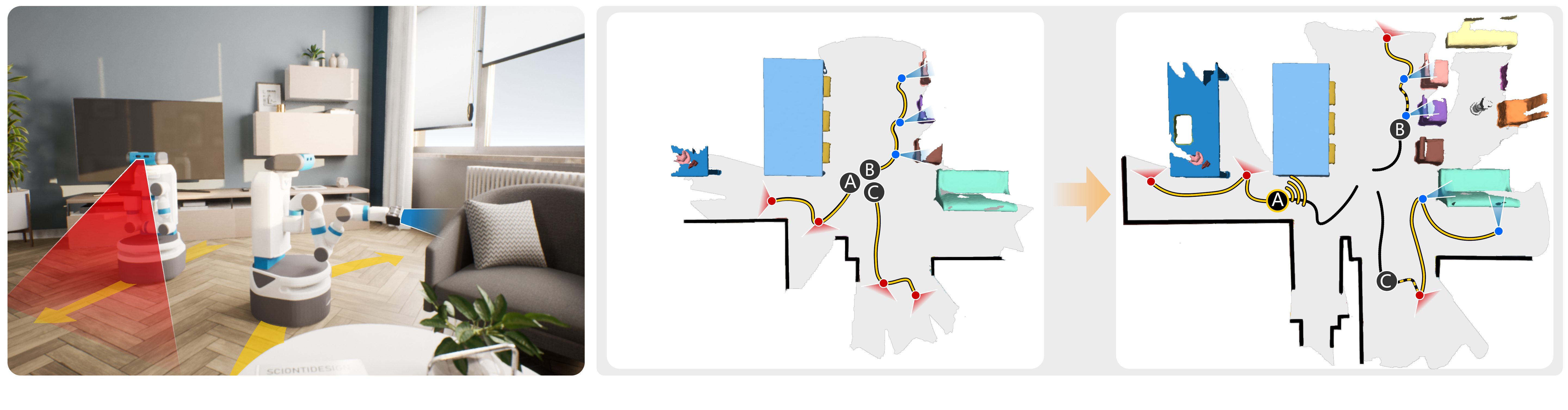}
        \put(2.2,0.3){\small (a) Two scanning modes: {\color{red} \textit{\rev{explorer}}} and {\color{blue} \textit{\rev{reconstructor}}}}
        \put(47,0.3){\small (b) Asynchronous collaborative autoscanning with mode switching}
    \end{overpic}
    \caption{ (a) Two scanning modes: \textit{\rev{explorer}} mode with rapid moving speed and far vision (shown in red) for  exploration task and  \textit{\rev{reconstructor}} mode with low moving speed and narrow vision (shown in blue) for object reconstruction task;
        (b) Our asynchronous collaborative autoscanning method: given the initially reconstructed scene by turning the three robots around their initial locations,
        our method first generates a set of tasks with the yellow path for each robot (left).
        Once one robot has completed all its assigned tasks,
        robot A in this case, new tasks will be generated and appended to all the robots immediately (right).
        The completed paths are shown in black, and the paths assigned in the previous round but haven't been completed are shown with dashed yellow lines on top of the black.
    }
    \label{teaser}
\end{teaserfigure}

\begin{abstract}
   When conducting autonomous scanning for the online reconstruction of unknown indoor environments, robots have to be competent at exploring scene structure and reconstructing objects with high quality.
Our key observation is that different tasks demand specialized scanning properties of robots: rapid moving speed and far vision for global exploration and slow moving speed and narrow vision for local object reconstruction,
which are referred as two different scanning modes: \textit{\rev{explorer}} and \textit{\rev{reconstructor}}, respectively.
When requiring multiple robots to collaborate for efficient exploration and fine-grained reconstruction, the questions on when to generate and how to assign those tasks should be carefully answered.
Therefore, we propose a novel asynchronous collaborative autoscanning method with mode switching, which generates two kinds of scanning tasks with associated scanning modes, i.e., exploration task with \textit{\rev{explorer}}  mode and reconstruction task with \textit{\rev{reconstructor}} mode, and assign them to the robots to execute in an asynchronous collaborative manner to highly boost the scanning efficiency and reconstruction quality.
The task assignment is optimized by solving a modified Multi-Depot Multiple Traveling Salesman Problem (MDMTSP).
Moreover, to further enhance the collaboration and increase the efficiency, we propose a \textit{task-flow} model that actives the task generation and assignment process immediately when any of the robots finish all its tasks with no need to wait for all other robots to complete the tasks assigned in the previous iteration.
Extensive experiments have been conducted to show the importance of each key component of our method and the superiority over previous methods in scanning efficiency and reconstruction quality.

%
%
%
%
%
%
%
%

\end{abstract}

\ccsdesc[100]{Computing methodologies~Shape analysis}

%
%

\keywords{Indoor scene reconstruction, autonomous reconstruction, multiple robots cooperation, asynchronous task assignment}
\maketitle

\section{Introduction}\label{introduction}
\rh{With an increasing demand for applications such as augmented and virtual reality, gaming and robotics,
    the research community has studied extensively toward generating digitized 3D indoor scenes with RGB-D sensors using robots
    ~\cite{charrow2015information,wu2014quality, xu2015autoscanning, huang2020autonomous}.
    To improve the scanning efficiency, there is a growing trend for adopting multiple robots to reconstruct the unknown area
    ~\cite{guan2006sensor, tian2021kimera, duhautbout2019distributed}.

    \rh{The work of \cite{dong2019multi} shows an impressive result in the field of multi-robot dense reconstruction of an unknown environment. }
    However, both scanning efficiency and reconstruction quality are sub-optimal as robots could remain \textit{inactivated} waiting for others before receiving new tasks synchronously, and objects may not be reconstructed with high quality due to rapid scanning without special focus.
    As noticed in some single-robot autoscanning systems \cite{xu2015autoscanning,xu20163d,huang2020autonomous,liu2018object},
    objects in the indoor scenes demand more careful scanning due to their fine-grained surface and self-occlusion.
    Therefore, how to deploy multiple robots to explore the unknown indoor scenes rapidly while obtaining reconstruction results with high completeness and accuracy
    is still an open question.

    %
    %


    %
    %
    %
}
\rh{
    %
    To realize high-quality reconstruction of the unknown indoor scenes with high efficiency,
    our key observation is that other than subdividing the mission into two tasks: exploration tasks and reconstruction tasks,
    specialized scanning properties of robots should be associated to different tasks to serve their goals better.
    The exploration task is designed to explore unknown regions and locate the targets for further reconstruction, which requires rapid moving speed and far vision,
    while the reconstruction task focuses on more detailed scanning of objects to obtain fine-grained geometries, which requires slow moving speed and narrow vision.
    We denote the above two different scanning modes as \textit{\rev{explorer}} and \textit{\rev{reconstructor}} modes, which are associated with the exploration and reconstruction tasks, respectively.
    To further require multiple robots to collaborate for high exploration efficiency and reconstruction,
    we propose a novel asynchronous collaborative autoscanning method with \textit{mode switching},
    which generates two kinds of scanning tasks with associated scanning modes, 
    and then assigns to the robots to execute in an asynchronous collaborative manner to highly boost the scanning efficiency and reconstruction quality,
    as shown in Figure \ref{teaser}.

    With the exploration tasks generated based on the current frontiers and reconstruction tasks aiming at incomplete objects,
    assigning these tasks to the robots according to the current situation is a complex problem.
    To achieve an optimal assignment with the least total traveling distance and processing time,
    we formulate the problem as a modified \textit{Multi-Depot Multiple Traveling Salesman Problem} (MDMTSP).
    However, reaching the exact solution of this problem consumes plenty of time and resources \cite{cheikhrouhou2021comprehensive},
    so it is impossible to search for the optimal solution when the assignment should process frequently.
    Therefore, we modify the multi-robot goal assignment solution \cite{faigl2012goal} to be suitable for our new problem formulation
    by approximating the optimal solution using an iterative optimization within a short time.
    %

    To prevent some robots from being inactive and waiting for incoming tasks,
    we propose a task-scheduling model named \textit{task-flow} to allow robots to receive new tasks asynchronously to further enhance the collaboration efficiency.
    Specifically, the  \textit{task-flow} model activates the control center to generate and assign new tasks when any of the robots finishes its current task sequence.
    As a result, all robots can be occupied with executing various tasks during the entire scanning procedure without waiting for others to finish their tasks.
    Both the decision-making and task-processing carry out asynchronously until the scene is fully explored with all objects carefully reconstructed.
    %
    %

    Our method has been implemented on top of Robot Operating System (ROS) \cite{ros} and tested using both multiple Fetch robots\cite{wise2016fetch} in simulation,
    and four \texttt{turtlebot3} robots \cite{turtlebot} in reality.
    Extensive experiments have been conducted to show the importance of each key component of our method and the superiority of our method over previous methods in scanning efficiency and reconstruction quality. 

    To summarize, we propose a novel asynchronous collaborative autoscanning method with mode switching, which efficiently utilizes multiple robots for exploring, reconstructing, and understanding an unknown scene.
    Our technical contributions include:
    \begin{itemize}
        \item  Two scanning modes tailored for the execution of two different scanning tasks, i.e., \textit{\rev{explorer}} for exploration task and \textit{\rev{reconstructor}} for reconstruction task.  %

        \item  A new modified MDMTSP and corresponding approximate solver to optimize each robot's task assignment and execution order for high efficiency and load balance.

        \item A task scheduling model named \textit{task-flow}  to minimize the waiting time of each robot and enable asynchronous collaboration among multiple robots.
    \end{itemize}
}

%
%
%

%

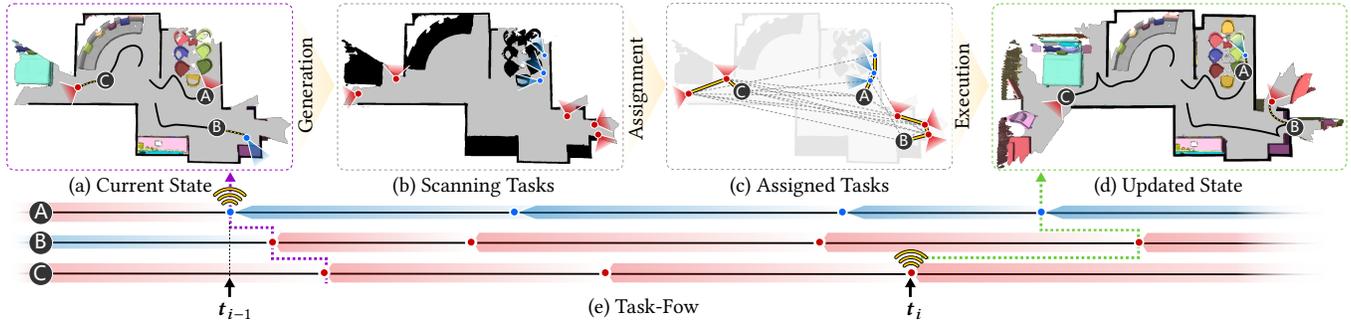
\begin{figure*}[t]
	\centering
	\begin{overpic}[width=1.0\textwidth,tics=5]{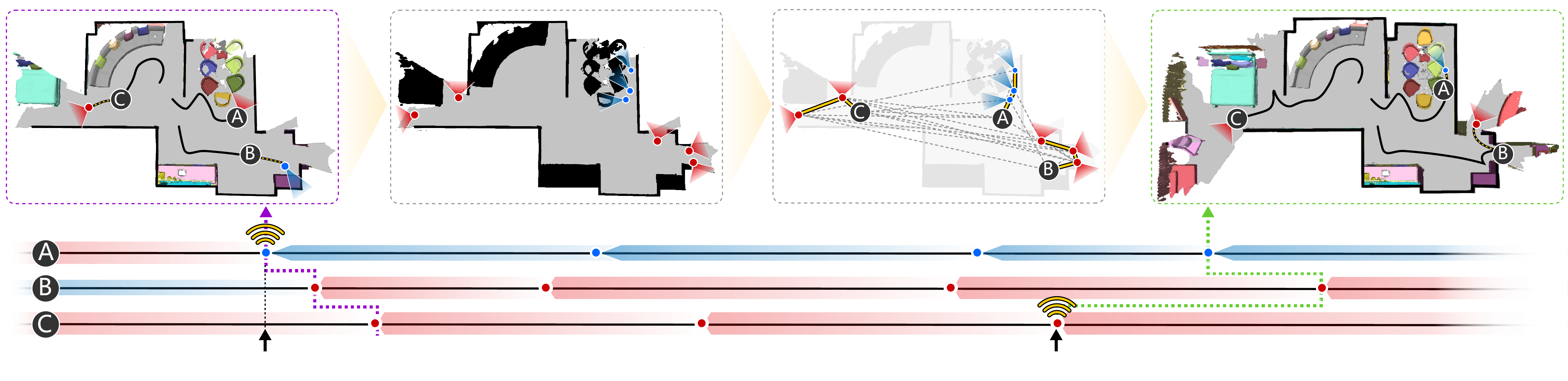}
		\put(5,9.7){\small (a) Current State}
		\put(29,9.7){\small (b) Scanning Tasks}
		\put(54,9.7){\small (c) Assigned Tasks}
		\put(81,9.7){\small (d) Updated State}
		\put(43.5,0.7){\small (e) Task-Fow}
    \put(22,14){\small \rotatebox{90}{Generation}}
    \put(46.5,13.5){\small \rotatebox{90}{Assignment}}
    \put(70.7,14.3){\small \rotatebox{90}{Execution}}
		\put(16,0.5){\small $\boldsymbol{t}_{i-1}$}
		\put(67,0.5){\small $\boldsymbol{t}_i$}
	\end{overpic}
	\caption{
	\rh{Overview of our asynchronous collaborative autoscanning method. Once a robot finishes its current tasks, robot A in this case (a),  new exploration tasks (red viewpoints) and reconstruction tasks (blue viewpoints) will be generated (b) and assigned to the robots by solving a modified Multi-Depot Multiple Traveling Salesman Problem (MDMTSP) (c).  The control center will be activated again if any of the robots finishes the new tasks again, robot C in this case (d).}
	}
	\label{fig:pipeline}
\end{figure*}

\section{Related work}
\subsection{Single-robot autonomous reconstruction}
To date, literature has dug deep into the field of autonomous reconstruction.
Studies devote to the scanning of a single object at first \cite{krainin2011autonomous,vasquez2014volumetric,wu2014quality},
then gradually move to the scanning of an entire scene \cite{ramanagopal2016motion,xu2015autoscanning,xu20163d,charrow2015information,schmid2020efficient}.
%
%
%
%
%
%
%
Unlike the works that focus on the \rev{quickly} covering of the unknown environment, there are also works trying to enhance the reconstruction quality simultaneously.
\rh{For example, Xu et al.~\shortcite{xu2017autonomous} harness a time-varying tensor field to plan the robot's movement and solve a 3D camera trajectory under the path constraint.
    The work of \cite{liu2018object} is the first one to utilize the semantic information of objects in path planning:
    they propose a \textit{next best object} (NBO) algorithm and a model-driven \textit{next best view} (NBV) algorithm to scan an indoor scene efficiently with high object reconstruction quality.
    A similar idea has also emerged in the field of reconstructing scenes with a single drone.
    Several methods \cite{heng2015efficient, DroneFly21, roberts2017submodular} are proposed to allow the (micro aerial vehicle) MAV
    to fulfill exploration and reconstruction objectives simultaneously by adopting an explore-then-exploit or a reconstruct-while-explore strategy.
    Deep learning-based methods \cite{sorokin2021learning,niroui2019deep,fang2019scene} also show promising results when building 2D maps of unknown scenes.
    Nonetheless, how to extend these approaches to scene-level 3D representation remains unsolved.}
%
%
%

Our work tries to achieve efficient exploration and high-quality reconstruction with a multi-robot system via using both geometry information of the environment and the semantic information of the objects.
Compared to a single robot, the multi-robot system has the advantage that it can process the exploration and reconstruction tasks in parallel instead of finding the balance of choosing between these two tasks and processing them sequentially.
However, the multi-robot system has to overcome the obstacle of adjusting the task assignment to save traveling energy and time.

\subsection{Multi-robot collaborative reconstruction}
%
Multi-robot systems can adopt either centralization or decentralization as their organization strategy.
Although the decentralized systems \cite{schneider1998territorial,sartoretti2019distributed,atanasov2015decentralized} have an unparalleled advantage in robustness,
the centralized system can improve cooperative efficiency by sharing global information \cite{matoui2020contribution,li2019robust} and unifying the decision-making process.
%
%
Since the communication and localization between robots and the control center can be solved by using auxiliary tools like motion capture cameras in indoor environments,
we utilize the centralized scheme in our work for a higher scanning efficiency.

Multi-robot active mapping has been an active research area for decades \cite{julia2012comparison,bhattacharya2014multi,burgard2005coordinated}.
Both rule-based \cite{faigl2012goal,albina2019hybrid} and learning-based strategies \cite{hu2020voronoi,ye2022multi} have been proposed to distribute the assignments among
robots to maximize the exploration coverage, but these methods still fail to produce a dense and high-quality reconstruction due to the lack of quality-driven view planning.
%
The most relevant work to ours is \cite{dong2019multi}.
They separate the scanning procedure into multiple intervals.
Each interval comprises a series of robot scanning, task extraction, task assignment, and path planning processes.
Despite the promising results, this method generates new reconstruction tasks only based on the frontiers without distinguishing the objects in the interior scenes.
As a result, some holes appear on the surface of the reconstruction results, leading to the declination of the reconstruction quality.
And this algorithm fails to take full advantage of each robot.
Since assigning the workload evenly is an NP-hard problem, the working time of each robot varies in each interval,
leading to the robots being inactivated and waiting for others.
In contrast, our method contains not only the exploration tasks to broaden the known area but also reconstruction tasks to scan the objects in detail.
We also add a \textit{task-flow} synchronization model to drive the robots and generate new tasks simultaneously,
no robots can be idle until the scene is fully reconstructed.

\subsection{Multi-Depot Multi-Travelling Salesman Problem}
Even though researchers have studied the optimal task assignment problem with constraints in engineering and economics from a variety of extents \cite{trigui2017fl,albina2019hybrid},
it is still a challenging research axis in the field of robotics.
The task assignment problem in a centralized multi-robot system is to drive multiple robots from various depots to multiple task positions with minimum travel distances.
The problem can be naturally formulated as a \textit{Multi-Depot Multi-Travelling Salesman Problem} (MDMTSP).
The work of \cite{sundar2017algorithms} introduces the integer linear programming formulation and proposes a branch-and-cut algorithm to reach the optimal solution.
This approach, along with other methods \cite{albina2019hybrid,vali2017constraint}, seek to solve the problem directly.
However, these solutions are \rev{too} time-consuming  \cite{cheikhrouhou2021comprehensive} to be adopted in the task assignment procedure.
There are also other approaches using the meta-heuristic algorithms \cite{yuan2013new,khoufi2015optimized, chen2018ant} to solve the problem from different aspects such as genetic algorithm or ant colony optimization.
Still, how to extend these methods to the new situation when there are two interchangeable identities in the formulation remains obscure.
In our approach, we extend the work in \cite{faigl2012goal} and modify it to approximate the exact solution via a divide-and-conquer strategy to reduce the computational complexity.


\section{Overview}\label{overview}
%


In this paper, we tackle the problem of automatic scene reconstruction with a centralized multi-robot system as in~\cite{dong2019multi}:
given an unknown indoor scene and the initial positions of several robots,
the goal is to explore and scan the scene such that the scanning coverage and reconstruction quality are maximized while the scanning effort is minimized.
Figure~\ref{fig:pipeline} shows an overview of one iteration of our asynchronous collaborative autoscanning method.

At the very beginning,  all the robots turn around to scan their surroundings so that an initial reconstruction is established to start the iteration.
Each time when a robot finishes its current task sequence,
the control center is activated to generate new scanning tasks based on the current reconstruction and assigns them to all the robots.
Then, each robot executes the assigned tasks with associated scanning modes to get the reconstruction updated.
\ok{To execute the new tasks assigned, each robot adapts its moving speed and camera view according to the task type. }
The control center will be activated again if any of the robots finish the new tasks again, and the iteration ends if no more new tasks can be generated.

%
%
%
%
\rh{\paragraph{Task generation}
    In our approach, the grand scanning task is divided into two sub-tasks:
    exploration tasks for high coverage and reconstruction tasks for high quality,
    which are represented using red viewpoints and blue viewpoints in Figure~\ref{fig:pipeline} (b), respectively.
    The exploration task is responsible for the rapid expansion of the known areas,
    and the corresponding viewpoints are generated based on the frontiers of the 2D occupancy grid updated with the 3D reconstruction results.
    The reconstruction task is responsible for covering the objects' surfaces with high scanning accuracy,
    and the corresponding viewpoints are generated based on the comparison between the predicted completion and the current scan.
}
\rh{More details about the task generation are provided in Section~\ref{sec:generate}.}



\rh{\paragraph{Task assignment}
    With the task viewpoints generated, we start optimizing the scanning paths for all robots. 
    We first construct a weighted graph by setting the task viewpoints and robot positions as nodes and their traveling paths as edges.
    To assign the generated tasks to the robots with minimum execution effort and time consumption,
    we then formulate the problem as a modified \textit{Multi-Depot Multiple Traveling Salesman Problem}  (MDMTSP) and search for the approximate optimal solution by an iterative optimization method.
    We determine the task distribution to the robots first. Next, the execution sequence of tasks for each robot is determined by solving a traditional traveling salesman problem (TSP).
    Examples of the constructed graph and the final assigned task sequences for all robots are shown in Figure~\ref{fig:pipeline}(c).
    More details about the formulation and solution of the task assignment can be found in Section~\ref{sec:assign}. }

%

\paragraph{Task execution}
Each robot maintains and moves along a path that connects a sequence of assigned task viewpoints to complete the tasks one by one, as shown in Figure~\ref{fig:pipeline}(d).
Meanwhile, the path keeps growing with new tasks assigned in each iteration.
Note that there are two kinds of tasks associated with two scanning modes. When the robot is processing one specific type of task, its moving speed and scanning range should be automatically shifted to the corresponding values.
In particular, the robot with \textit{\rev{explorer}} mode processes the exploration task by tilting the head up and down at a fixed height to scan forward \ok{when arriving at the task position}.
With the known areas expanded, objects are detected by \textit{Mask R-CNN} ~\cite{he2017mask} and the semantic information is stored in the reconstruction result.
\ok{To enlarge the known area, robots keep scanning on the route. The intermediate scans along the task path are also utilized.}
For reconstruction tasks, the robot with \textit{\rev{reconstructor}} mode uses the camera holding in hand with a 7-DOF arm to scan the objects carefully and stably with a particular angle and distance.
To achieve a high-quality reconstruction, the robot moves continuously and meticulously to ensure complete capture and sufficient overlap between adjacent scanning frames to reduce the error in object reconstruction.
Figure \ref{teaser} gives an illustration of these two different scanning modes when executing the corresponding tasks.


\rh{\paragraph{Task scheduling.}\label{asy-section}
    Due to various reasons, including different task difficulties, it is often irrational to expect all robots to finish their tasks simultaneously,
    Therefore, a less time-consuming way is to let the robot automatically carry out new tasks after completing its assigned tasks in the last round.
    Under this strategy, we propose the \textit{task-flow} model to accelerate the task scheduling process, as shown in Figure~\ref{fig:pipeline}(e).
    Once a robot is about to finish its current tasks, it signals the control center to trigger the decision-making stage.
    After analyzing the geometry and semantics of the current scene, the control center starts a new iteration of task generation and assignment to append new tasks to the end of each robot's task path.
    As a result, while all robots are processing the stored tasks, all the task paths keep updated and no robot needs to wait for other robots,
    which highly enhances the system's efficiency.
    While the scanning procedure runs, tasks are generated from the control center and processed by robots like a flow.
    This procedure runs continuously until no latest task is updated, which indicates that the entire scanning task has been completed.
}

\begin{figure}[t]
	\centering
	\begin{overpic}[width=1.0\linewidth,tics=5]{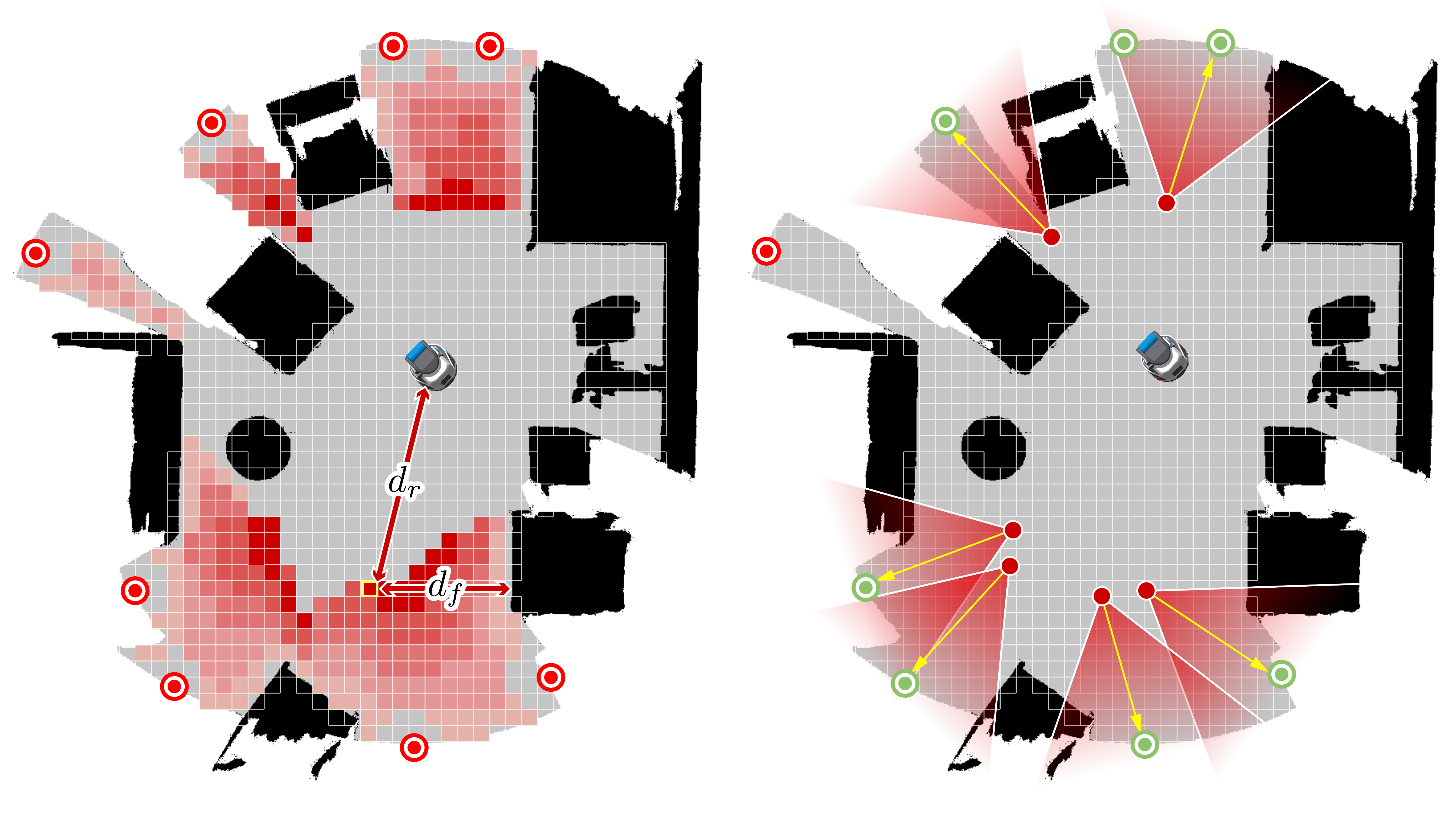}
		\put(15,0.7){\small (a) Validity score}
		\put(65,0.7){\small (b) Selected points}
	\end{overpic}
	\caption{
  	\rh{
	Exploration task generation with viewpoint selection on 2D occupancy grid. 
	(a) A set of frontiers (red circles) are first selected to determine the corresponding candidate viewpoints, and the validity score of each candidate viewpoint is defined by measuring how close it is to the nearest robot ($d_r$) and how far from the nearest obstacle ($d_f$) . 
	(b) A set of viewpoints (red dots) with high validity scores are selected with the constraint that each frontier can only be covered once. Then frontiers covered by the field of vision of the selected viewpoints are turned into green.
	}
	}
	\label{fig:frontier-tasks}
\end{figure}

\section{Task generation}
\label{sec:generate}

\rh{
\subsection{Exploration task generation}
Given $R$ robots $\{\mathcal{R}_r\}_{r=1}^{R}$ in an unknown environment,
we \rev{represent} each robot $\mathcal{R}_r$ with a tuple $\mathcal{R}_r=\{\boldsymbol{r}_r, \theta_r, \mathit{I}_r\}$.
The tuple contains the elements of the robot's position and orientation on the 2D occupancy grid as well as the scanning mode.
The exploration task is also defined on the 2D map and denoted as $T_i^{exp}=(x_i, y_i, \theta_i)$, encoding its position and orientation.

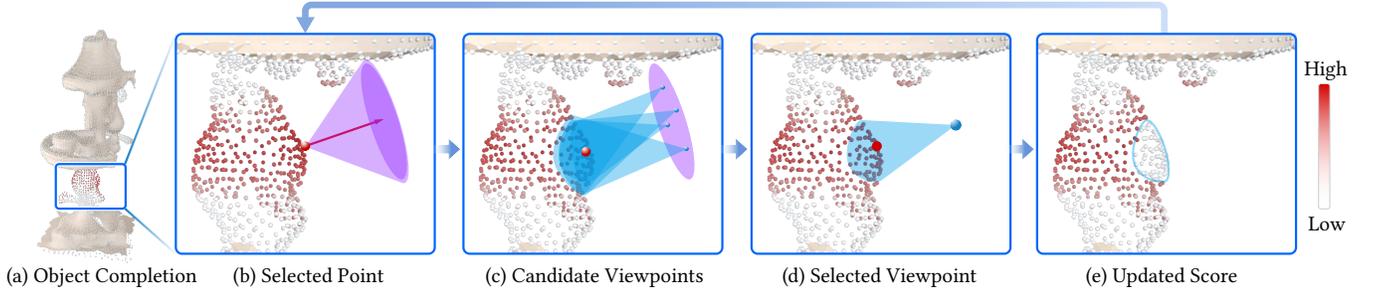
\begin{figure*}[t]
	\centering
	\begin{overpic}[width=\linewidth,tics=5]{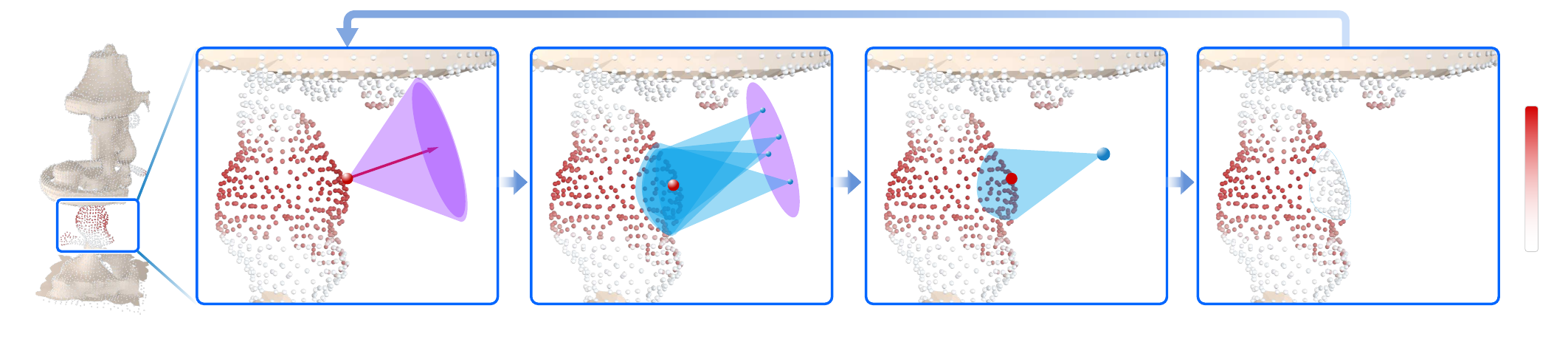}
		\put(0,0.7){\small (a) Object Completion}
		\put(16.8,0.7){\small (b) Selected Point}
		\put(35.5,0.7){\small (c) Candidate Viewpoints}
		\put(57.5,0.7){\small (d) Selected Viewpoint}
		\put(80,0.7){\small (e) Updated Score}
		\put(96.5,4.5){\small Low}
		\put(96.2,16){\small High}
	\end{overpic}
	\caption{
  	\rh{
	Reconstruction task generation based on the completeness analysis on the 3D point cloud. 
	(a) Object scan with the completed point cloud, and the color on the point indicates the corresponding incompleteness score.
	(b) The point with the highest incompleteness score is selected, and an optic cone is created for candidate viewpoints selection.
	(c) An optic cone is further created for each candidate viewpoint to compute its vision coverage. 
	(d) The viewpoint with the highest vision coverage is selected.
	(e) The incompleteness score is updated by removing the region covered by the selected viewpoint.
	}
	}
	\label{fig:reco-tasks}
\end{figure*}

To better explore the unknown area, we select the viewpoints for exploration tasks based on the frontiers of the current 2D occupancy grid,
as shown in Figure \ref{fig:frontier-tasks}.
In detail, we first extract all frontiers via the Canny algorithm \cite{canny1986computational} and then perform farthest sampling to get the set of uniformly distributed frontiers for further exploration,
shown as the red points in Figure \ref{fig:frontier-tasks} (a).
Then, to have a better vision for each frontier, we restrict the scanning distance to be within a predefined range and further exclude all the grids inside the region where the frontier point is not visible.
As a result, only a subset of grids on the 2D map become candidate viewpoints.
As later the robots need to move to the selected viewpoint to conduct the scan,
we expect the viewpoint to be close to the nearest robot to minimize the moving effort as well as far from the obstacles for safety.
So we further define a validity score on each candidate viewpoint as $s  = d_r - d_f$,
where $d_r$ is the distance to the nearest robot and $d_f$ is the distance to the nearest unfree grid,
which are visualized using the colormap shown on the grids in Figure \ref{fig:frontier-tasks} (a).

With the candidate viewpoints with computed validity scores, we select the final viewpoints iteratively.
Specifically, we first select the viewpoint with the highest validity score and then find the frontiers that can be covered by this viewpoint when setting the vision angle to be in the \textit{\rev{explorer}} mode.
Then, the validity scores will be updated by setting the score of any candidate viewpoint associated with the covered frontiers to be zero to avoid further selection.
The iteration continues by selecting the next viewpoint with the highest score until a pre-defined number of viewpoints have been selected.
Figure \ref{fig:frontier-tasks} (b) shows the selected viewpoints and the frontiers covered by their visions turn green.
}

\subsection{Reconstruction task generation} \label{reconstruction task generation}
When reconstructing objects, the viewpoint of the sensor need to be placed in 3D space to scan the objects from various views,
as the reconstruction quality can be quite limited if the camera height is fixed.
Therefore, the reconstruction task is denoted as $T_i^{rec} = (x_i, y_i, z_i, \theta_i, \phi_i)$,
where $\theta_i\in[0,2\pi)$ is the orientation,
and $\phi_i \in [-\frac{\pi}{2}, \frac{\pi}{2}]$ is the elevator angle of the view point.

As the goal of the reconstruction task is to get high-quality reconstruction of the 3D objects, more scans should be conducted around the incomplete objects, especially toward regions with big holes.
Thus, we keep track of the object instances during the whole scanning process as in the work of \textit{Voxblox++} \cite{grinvald2019volumetric} and analyze the completeness of each object for viewpoint selection.
\ok{Note that by using a combined geometric-semantic segmentation scheme, the mapping framework \textit{Voxblox++} is able to detect recognized elements from a set of known categories and simultaneously discover previously unseen objects in the scene. Thus, we can detect almost all the objects in the scene which have been scanned. }

Given a reconstructed object with a unique instance label,
we first uniformly sample  $N = 2048 $ points $\mathcal{P}$ on the object surface and  use the GR-Net~\cite{xie2020grnet} to predict the corresponding complete point cloud $\mathcal{C}$.
Then, the incompleteness score of each point $\boldsymbol{q}\in\mathcal{C}$ is defined as its nearest distance to the original point cloud $\mathcal{P}$ :
\begin{equation}
    \label{point-completeness-error}
    S_c(\boldsymbol{q}) = \mathop{\min}\limits_{\boldsymbol{p}\in\mathcal{P}}||\boldsymbol{q}-\boldsymbol{p}||_2,
\end{equation}
which is then further normalized by  the maximal distance among the set: $\bar{S}_c(\boldsymbol{q})  = S_c(\boldsymbol{q})  / \mathop{\max}\limits_{\boldsymbol{q}\in\mathcal{C} }S_c(\boldsymbol{q}) $.
\ok{It is worth noting that, the completion process is only used to search for the incomplete parts of the objects and will not be used directly as the reconstruction results, so the prediction doesn't have to be very precise.}

To filter out the objects that have already been completely reconstructed,
we compute the average incompleteness score $\bar{S}_c(\mathcal{C})$ of all points belonging to the complete point cloud $\mathcal{C}$,
and set $\tau=0.2$ as a heuristic threshold that if $\bar{S}_c(\mathcal{C})$ is less than $\tau$, we consider the object to be complete and no further scanning is needed.
For the remaining incomplete objects, we rank them based on the $\bar{S}_c(\mathcal{C})$ score and select the viewpoints one by one.
For the selected incomplete object $\mathcal{C}$,
we iteratively select a set of viewpoints around the shape based on the incompleteness score defined on points to make full coverage of the incomplete region as shown in Figure \ref{fig:reco-tasks}.

In more detail, \rev{to search for viewpoints with large coverage of incomplete regions, }
the point $\boldsymbol{q}$ with the highest incompleteness score will be selected first to generate a set of candidate viewpoints $\{v_i\}$ on the bottom surface of an optic cone with $\boldsymbol{q}$ as the apex and its normal direction $\mathbf{n}$ as the rotation axis, and then another optic cone with each reachable candidate viewpoint $v_i$ pointing to $\boldsymbol{q}$ are created to calculate its coverage,
as shown in Figure \ref{fig:reco-tasks}(b) and (c).

The view coverage of each viewpoint $v_i$ is defined by the sum of incompleteness scores of all the points on $\mathcal{C}$ that are visible within the optic cone of $v_i$ \rev{as well as within the reachable range of the robot's arm}.
Note that the vision range of the optic cone of each candidate viewpoint is set to be the same as that of the \textit{\rev{reconstructor}} mode,
while the vision range of the optic cone of $\boldsymbol{q}$  is doubled.
Once the viewpoint with the highest view coverage is selected,
the incompleteness scores will be updated for next viewpoint selection, as shown in Figure \ref{fig:reco-tasks}(c) and (d).

Note that, in practice, it \rev{often} happens that the object cannot be completed due to the unreachable holes on the reconstruction surface and scanning only once cannot complete the object to an impressive level.
Therefore, we set a historical record for all selected viewpoints of each recognized object.
If an object has been scanned three times along neighboring viewpoints without increasing the surface completeness,
no more reconstruction tasks will be generated inside the neighborhood of those viewpoints to avoid time-wasting in circling the "impossible missions".

\rh{
\section{Task assignment}
\label{sec:assign}
\subsection{Problem formulation}
Once the new exploration tasks $\mathcal{T}^{exp}$ and reconstruction tasks $\mathcal{T}^{rec}$ are generated, they need to be distributed to the robots for execution.
Note that due to our asynchronous \textit{task-flow} model, some of the robots may still have unfinished tasks assigned in the last round,
so the newly assigned tasks should be appended to the end of the current paths.
For robot $\mathcal{R}_r$,  we denote its unfinished task sequence as $\mathcal{T}^{rest}_r$ \ok{ and the last unfinished task as $T^{end}_r$.}
To find the best task assignment, we first construct a weighted graph $\mathbf{G} = (\mathcal{V}, \mathcal{E})$ to encode the spatial relationship between the tasks and robots,
where $\mathcal{V} = \mathcal{T}^{exp}\cup \mathcal{T}^{rec} \cup \mathcal{T}^{end} $ consists of all new tasks and the set of final tasks of all the robots in last round  $\mathcal{T}^{end} = \{T^{end}_r\}_{r=1}^R$,
and $\mathcal{E}$ consists of edges connecting each pair of nodes with distance calculated using A* algorithm \cite{hart1968formal}.
\ok{As the processing time of each task can be considered as fixed during the execution, we focus on optimization of the traveling cost during the task assignment and formulate it as a modified MDMTSP,
where the goal is to find a set of disjoint paths $\{\mathcal{T}_r\}_{r=1}^R$ that covers the whole set $\mathbf{G}$ such that the sum of all robots' tour costs is minimized:}
\begin{equation}
    {E}_{d}=\sum_{r=1}^{R} \left(   \sum_{ T_k\in\mathcal{T}_r } d(T_k, T_{k+1})  \right),
\end{equation}

However, extra demands and constraints make our problem formulation slightly different.
As our method focuses on multi-robot collaboration, other than minimizing the total tour cost, we also attempt to achieve load balance among robots to reduce the total time consumption of the assigned tasks.
Thus, we dispatch the robots with uniform workload to decrease this time consumption by including the capacity term:
\begin{equation}
    \label{eq:cap}
    {E}_{c}=\sum_{r=1}^{R} \left( (|\mathcal{T}_r| + | \mathcal{T}^{rest}_r|-C_r)^2 \right),
\end{equation}
where $ C_r=\mathop{\sum}_{r=1}^R ( |\mathcal{T}_r| +  |\mathcal{T}^{rest}_r|  ) /R$
\ok{is the total remianing tasks divided by the number of robots, standing for the average task capacity to each robot.}
As a result, our problem becomes a modified MDMTSP with the goal to minimize the total energy cost: 
\begin{equation}
    \mathcal{T}^{*} = \mathop{\arg\min}\limits_{\mathcal{T}=\{\mathcal{T}_r\}_{r=1}^{R}}  {E}_{d}+\ {E}_{c}.
    \label{terms-formular}
\end{equation}

Moreover, different types of tasks are associated with different scanning modes, which means that the robot needs to switch the mode when starting a different kind of task, leading to extra costs.
Besides, since reconstruction tasks aim to obtain fine-grained geometry of 3D objects, it is often preferable to have a sequence of consecutive careful scans.
Therefore, to keep the continuity of high-quality scanning toward the object and reduce the times of mode shifting,
we add an extra constraint to our objective function so that each task sequence $\mathcal{T}_r$ only contains one type of task:
$$
    \begin{aligned}
         & \prod_{T_k\in\mathcal{T}_r}\mathcal{I}(T_k)+\prod_{T_k\in\mathcal{T}_r}  \left(1-\mathcal{I}(T_k) \right)=1. \\
    \end{aligned}
$$
where $\mathcal{I}$ indicates the scanning mode corresponding to the task, and is set to be $1$ for \textit{\rev{explorer}} and $0$ for \textit{\rev{reconstructor}}.
}

\begin{algorithm}[t]
\caption{Updated MDMTSP problem solver}\label{algo:mdmtsp}
\KwIn{
 Current weighted graph $\mathbf{G}$\;
 Robot positions $\mathcal{R}$\;
 Exploration task positions $\mathcal{T}^{Exp}$\;
 Reconstruction task positions $\mathcal{T}^{Rec}$\; }
 \KwOut{Path node sequence for each robot:
   $$\mathcal{S}=\bigg\{..., \{\mathcal{R}_r, \mathcal{T}^{rest}_r, T_{r_1}, T_{r_2},...\},...\bigg\}$$}
 $\gamma \gets$ \texttt{InitialClustering}$(\mathcal{R}, \mathcal{T}, \mathbf{G})$\;
 $\gamma \gets$ \texttt{SimulatedAnnealing}$(\gamma, \mathbf{G})$\;
\For {$\gamma_i\in \gamma$}{
   $\mathcal{S}_i=$\texttt{TSPSolver}$(\gamma_i, \mathbf{G})$}
 $\mathcal{S} \gets \{ \mathcal{S}_1,...,\mathcal{S}_R\}$;
\end{algorithm}

\subsection{Modified MDMTSP solver}\label{MDMTSP solver}
It is proved that MDMTSP is NP-hard \cite{yadlapalli2009lagrangian}, and with extra constraints and demands in our formulation,
it is almost impossible to find an exact optimal solution within a short time.
To overcome the problem of computational complexity, we approximate the solution to this complex problem by splitting the model into three parts:
assigning scanning modes to robots, clustering tasks for each robot, and planning task execution order inside each cluster.
Since these three parts are inseparably interconnected, we propose an iterative approach to this chicken-and-egg situation.

The pseudocode of our solution is provided in Algorithm \ref{algo:mdmtsp}.
The method starts with an initial task clustering based on a heuristic method.
Specifically, we first compute the proportion of two types of viewpoints and randomly assign the scanning modes to robots according to this proportion with a guarantee that each scanning mode is assigned to at least one robot.
Then we cluster the tasks using the k-means clustering algorithm with the robots' end positions $\mathcal{T}^{end}$ as initial centroids and assign tasks to the robots with the same type and the least traveling cost. 
Given this initialization, the clustering is iteratively optimized through a simulated annealing algorithm.
We use the method of cluster-and-assign similar to \cite{faigl2012goal} in each iteration of simulated annealing,
and in each annealing iteration, two types of disturbance can be chosen randomly:
1) \textit{Exchange}: we separately select two robots from the \rev{explorer}s and the \rev{reconstructor}s and switch their scanning modes;
2) \rev{\textit{Reassign}}: we randomly change the scanning mode of a robot in the category containing more robots.
Once the clusters are optimized,  we deploy a traditional TSP solver to further optimize the execution sequence of tasks within each cluster.
As a result, each robot receives a new task sequence and appends them to the current workload.

\paragraph{Traveling distance term}
During the clustering optimization, it is required to calculate the traveling distance of each robot for finishing all the assigned tasks, which is computationally inefficient.
Accordingly, we use the method similar to \cite{dong2019multi} to approximate the traveling distance to decrease the computation cost.
More specifically, to estimate the distance from the robot to the assigned tasks,
we sum up the traveling distance between the tasks and \rev{their} centroid and the distance from the centroid to the robot's end position $\mathcal{T}^{end}_r$.
Note that different tasks are associated with different scanning modes at different speeds, affecting the assigned tasks' traveling time.
So when considering the scanning speed of different tasks $v(T)$, we denote the approximate traveling energy as:
\begin{equation}
    {E}_{d}^{'}=\sum_{r=1}^{R}\bigg(\sum_{T_k\in\mathcal{T}_r}\frac{d(T_k,\omega_r)}{v(T_k)}+ \frac{d({T}^{end}_r, \omega_r)}{v({T}^{end}_r)} \bigg),
\end{equation}
where $\omega_r$ is the centroid of the tasks cluster assigned to robot $\mathcal{R}_r$,
and $d(\cdot, \cdot)$ represents the traveling distance.

To combine the capacity term ${E}_{c}$ defined in Equation \ref{eq:cap} into the distance of the clustering method,
we perform a soft clustering method with modified \textit{Gauss Mixture Model} (GMM),
where the capacity term is multiplied by each robot's likelihood to consider the unprocessed tasks.
Unlike other clustering methods such as the k-means, it is more convenient for \textit{GMM} to add terms like capacity uniformity to the optimization process.
After each optimization step, we calculate the assignment cost of the new cluster and decide whether to update the states.

\begin{figure}[t]
  \centering
  \includegraphics[width= 0.8\linewidth]{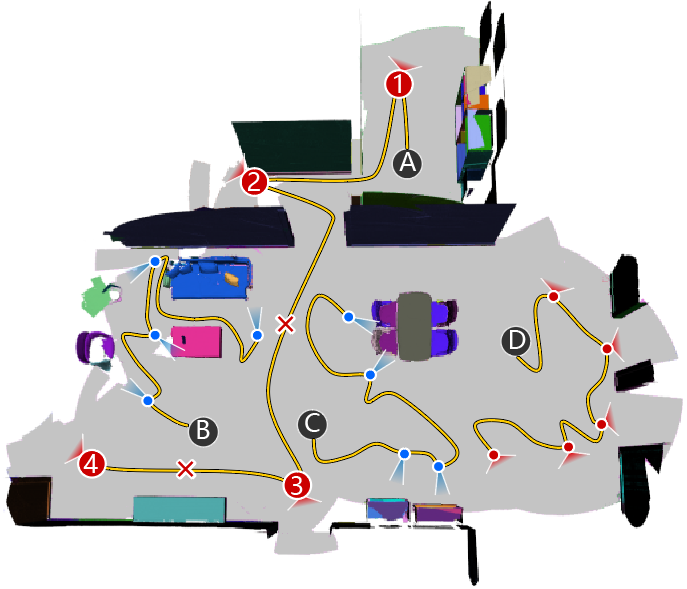}
  \caption{
  	\rh{
   Illustration of traveling energy constraint.
    When assigned with a task that has a long traveling distance, e.g., the task 3 for robot A in this case, 
    the scanning efficiency can be significantly reduced, thus all the subsequent tasks of task 3, including task 4, are deleted together for robot A, and it will wait for the task assignment in next iteration.
}
    }
  \label{fig:assign-constraint}
\end{figure}
\paragraph{Traveling energy constraint}
As the robot A shown in Figure \ref{fig:assign-constraint},
there are cases when the robot has no choice but to travel a long distance within one cluster to process tasks in the distance.
In practice, a better choice is usually to stay and wait for the tasks generated nearby in the next iteration.
The appearance of this inefficient situation is likely related to the fact that robots are ordered to process all generated tasks.
However, there are also other situations where long-distance traveling is necessary when no possible tasks remain close to the robot.
So to achieve a balance, we set a maximum travel distance between two consecutive tasks in a task sequence, and only check the task distance starting from a pre-defined index.
Once we find that one task's travel distance is larger than the given threshold, we remove all its subsequent tasks together to facilitate the execution. Thus this robot is more likely to finish all its remaining tasks earlier than other robots to trigger the control center to assign more practical tasks.
Experiments on this traveling energy constraint can be found in the supplementary material.




\section{Results and evaluation}
\subsection{System and dataset} \label{robot-system}

\paragraph{Robot system}
Our system runs on top of ROS, which supports the robot's standard behaviors, such as navigation and arm actions.
For the convenience of development and evaluation, we use the Gazebo \cite{gazebo} framework as our simulator, which can simulate the interactions with real-world. 
We further add synthetic noise to the depth maps to simulate the camera noise in reality, as in the work of \cite{handa2014benchmark}.
We use \texttt{Fetch} as our robots in the simulator.
Each robot is equipped with a \texttt{Primesense Carmine 1.09} mounted on the top of it, a \texttt{RealSense Depth Camera D435i} held in hand, as well as a \texttt{SICK 2D sensor}.
The \texttt{SICK} is used for robot tracking, which is necessary for exploration tasks and map merging.
A desktop PC processes all data with Intel(R) Core(TM) I7-10700K CPU (3.8GHz$\times$8), 64GB RAM, and an Nvidia GeForce RTX3080 GPU.
The GPU is used for shape completion (GR-Net) and instance segmentation (Mask R-CNN)\ok{, which takes about 16GB RAM and 10GB GPU memory in total}.
\ok{The RGB-D data and camera-pos captured by all robots are utilized in our methods. }

For experiments in reality, we use \texttt{Turtlebots3} instead due to its lower price.
Each robot is equipped with a \texttt{RealSense Depth Camera D435i},
as well as an \texttt{RPLIDAR A2M8 $360^\circ$} LiDAR scanner and \texttt{Nokov} motion capture cameras (https://www.nokov.com/).
When the robots move beyond the capture range of the \texttt{Nokov}, the LiDAR scanners can track the robots' locations as well.

\paragraph{Scene dataset}

We select virtual scenes from two datasets:
synthetic scenes from \textit{Front3D} \cite{front3d}, which are composed of hand-made residential buildings delineated by clean boundaries and separations,
and realistic scenes from \textit{Matterport3D} \cite{matterport3d}, which are reconstructed from real scenes and contain more types of indoor scenes such as offices and hotels.
The collection contains 40 scenes including residential buildings (30), offices (5), restaurants (2), and other environments (3).
All these indoor scenes can be regarded as \textit{flat-layout} of walls and furniture, without stairs or sunken regions.
To avoid wasting time in meaningless exploration beyond the scene, we seal off all windows and doors in the scenes.
In our evaluation, we carefully select four representative benchmark scenes in each category of scales: small ($< 230 m^2$),  medium (230-300 $m^2$), and large ($>300 m^2$) in our dataset to frequently be used in most of the quantitative plots in this section.

\rh{
    \subsection{Evaluation metrics}
    As our goal is to conduct automatic scene reconstruction with multi-robot collaboration, we evaluate the method in three aspects: reconstruction quality, execution efficiency, and load balance among multiple robots.


    \paragraph{Reconstruction quality}



    We evaluate how complete and accurate the objects with detailed geometry are captured.

    \textbf{Object Completeness (O-Comp)} is measured by the average percentage of the covered surface of all objects in the scene, and the completeness of the reconstructed point cloud $\mathcal{O}$ compared to the ground truth point cloud $\mathcal{G}$ is defined as:
    \begin{equation}
        OC (\mathcal{O}, \mathcal{G})=\frac{1}{|\mathcal{G}|}\mathop{\sum}\limits_{\boldsymbol{q}\in{\mathcal{G}}} \mathbbm{1} \big(d(\boldsymbol{q}, \mathcal{O}) \le \tau\big),
        \label{comp}
    \end{equation}
    where $d(\boldsymbol{q}, \mathcal{O}) $ is the minimal distance of the point $\boldsymbol{q}$ to the point cloud $\mathcal{O}$,
    and $\tau$ is the predefined distance threshold. 

    \textbf{Object Accuracy (O-RMS)} is measured by the average distance of the reconstructed objects in the scene,
    where the distance between each reconstructed point cloud $\mathcal{O}$  and the ground truth point cloud $\mathcal{G}$ is defined as:
    \begin{equation}
        OA(\mathcal{O},\mathcal{G})=\frac{1}{|\mathcal{O}|}\mathop{\sum}\limits_{\boldsymbol{p}\in{\mathcal{O}}}\mathop{\min}\limits_{\boldsymbol{q}\in\mathcal{G}}||\boldsymbol{p}-\boldsymbol{q}||_2.
        \label{rms_acc}
    \end{equation}

    \paragraph{Execution efficiency}
    To measure scanning efficiency, we compute the \textbf{Distance Consumption (D-C)}
    for the total travel distance of all robots and the \textbf{Time Consumption (T-C)}
    for the total time consumed in the scanning procedure.

    \paragraph{Load balance.}
    To measure the \textbf{Distance Load Balance(D-LB)}, we use the method in the work of \cite{dong2019multi} and compute the ratio of the standard deviation and mean value corresponding to the moving distance of all robots.
    Moreover, we further compute the \textbf{Time Load Balance(T-LB)}, which is the percentage of the waiting time  of all robots during the scanning process.}

\subsection{Comparison to baselines}
\rh{\paragraph{Baseline methods.}
    To the best of our knowledge, the work of \cite{dong2019multi} is the only work that shares the same goal as our work, so it's a natural baseline to compare with.

    To further evaluate the effect of multi-robot cooperation compared to isolated robots, we also compare our method with the method proposed in \cite{liu2018object}, the state-of-the-art single-robot reconstruction work, used in two different settings.
    The first is the original single-robot setting, denoted as NBO $\times$ 1, where evaluation metrics like time consumption and distance consumption of this baseline will be divided by the number of robots, 4 in our experiment, for a fair comparison. 
    The second setting, denoted as NBO $\times$ 4, assumes that we have extra prior knowledge of the scene structure and can subdivide it into isolated sub-regions for different robots to explore individually. In this case, each robot maintains its own reconstruction result and occupancy map, and no message will be exchanged during the scanning.
    \ok{Moreover, all baselines are modified by using the Voxblox++  as the underlying reconstruction module as in our method for fair comparisons. }

}

\begin{table}[t]
\centering
\caption{\rh{Comparison with the work of \cite{dong2019multi} on reconstruction quality of objects, reconstruction efficiency, and load balance.}}
\label{tab:compare-to-dong}
\begin{tabular}{ccccccc}
\toprule
\multirow{2}{*}{Quality} & \multicolumn{3}{c}{Object Completeness} & \multicolumn{3}{c}{Object Accuracy}
\\ 
\cmidrule(r){2-4} \cmidrule(r){5-7}
& Small & Medium & Large & Small & Medium & Large
\\ 
\hline
Dong & 53.85 & 41.49 & 40.27 & 0.062 & 0.081 & 0.097
\\
 \bf{Ours} & \bf{66.18} & \bf{72.49} & \bf{70.03} & \bf{0.035} & \bf{0.039} & \bf{0.033}
\\
\bottomrule
\toprule
\multirow{2}{*}{Efficiency} & \multicolumn{3}{c}{Time Consumption} & \multicolumn{3}{c}{Distance Consumption}
\\ 
\cmidrule(r){2-4} \cmidrule(r){5-7}
& Small & Medium & Large & Small & Medium & Large
\\ 
\hline
Dong & 19.3 & 23.0 & 28.8 & \bf{482.6} & \bf{525.8} & \bf{738.6}
\\
 \bf{Ours} & \bf{14.0} & \bf{18.9} & \bf{24.7} & 536.1 & 620.7 & 848.5
\\
\bottomrule
\toprule
\multirow{2}{*}{Balance} & \multicolumn{3}{c}{Distance Load Balance} & \multicolumn{3}{c}{Time Load Balance}
\\ 
\cmidrule(r){2-4} \cmidrule(r){5-7}
& Small & Medium & Large & Small & Medium & Large
\\ 
\hline
Dong & 0.172 & 0.216 & 0.159 & 0.205 & 0.262 & 0.273
\\
 \bf{Ours}& \bf{0.151} & \bf{0.200} & \bf{0.157} & \bf{0.062} & \bf{0.082} & \bf{0.093}
\\
\bottomrule
\end{tabular}
\end{table}

\begin{figure}[t]
	\centering
	\begin{overpic}[width=\linewidth,tics=5]{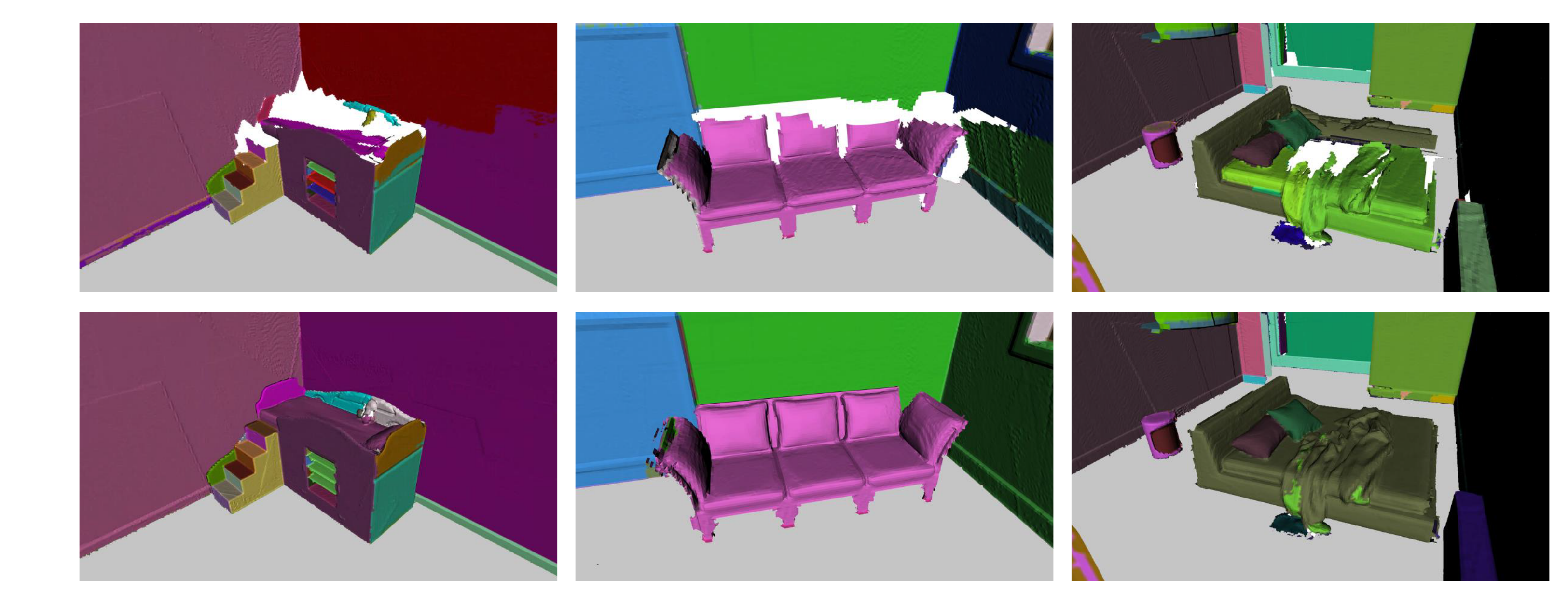}
		\put(1,24.5){\small \rotatebox{90}{Dong}}
		\put(1,6.5){\small \rotatebox{90}{Ours}}
	\end{overpic}
	\caption{
		\rh{Some of the intermediate results of \cite{dong2019multi} (top) comparing to ours (bottom).
		}}
	\label{fig:compare-to-dong-object-quality}
\end{figure}

\begin{figure*}[t]
	\centering
	\includegraphics[width=\linewidth]{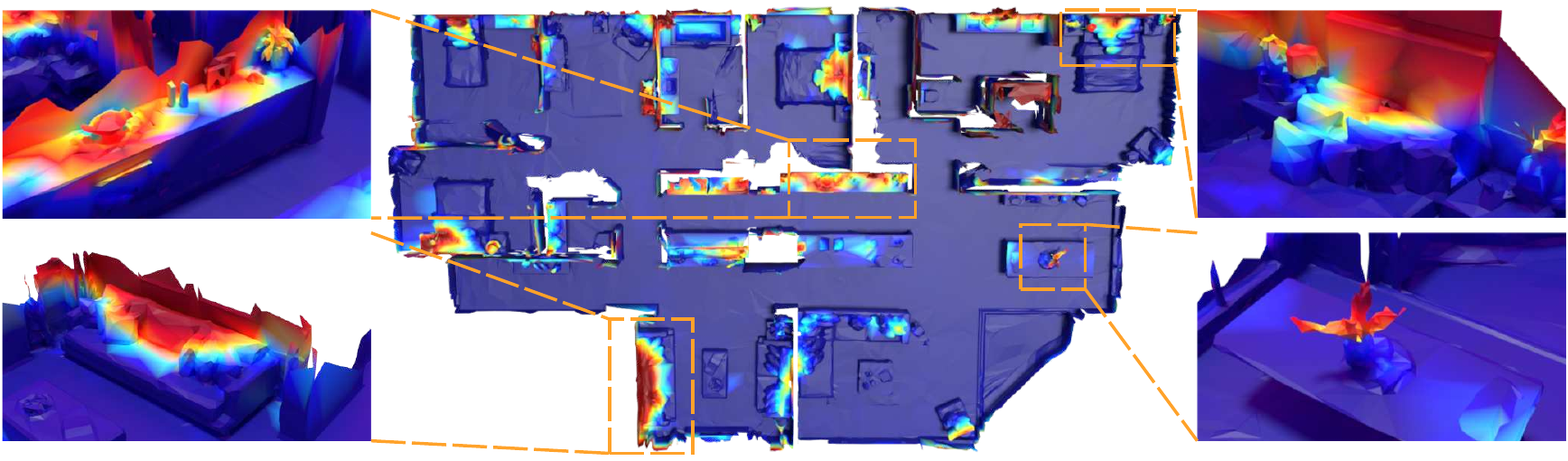}
	\caption{
		\ok{ Reconstruction error difference between the method of \cite{dong2019multi} and our method. Red indicates a higher reconstruction error of \cite{dong2019multi} compared to ours while blue indicates similar reconstruction quality.  }
	}
	\label{fig:compare-to-dong-colormap}
\end{figure*}

\paragraph{Results}
Table \ref{tab:compare-to-dong} shows the comparisons of objects' reconstruction quality, reconstruction efficiency,
and load balance between Dong and our method.
The metrics show that our method surpasses the work of \cite{dong2019multi}  in object completeness and accuracy,
which justifies the use of reconstruction tasks associated with tailored scanning mode in our method.
Figure \ref{fig:compare-to-dong-object-quality} shows some visual comparisons on the reconstruction results.
\ok{Due to the nature of their method~\cite{dong2019multi}, no more task viewpoints will be generated surrounding the incomplete objects, and thus these incomplete results remain until the end of the mission. We also compare the accuracy error between ours and \cite{dong2019multi} by calculating the relative difference per point to the ground truth $\mathcal{G}$.
    \begin{equation*}
        AE(\boldsymbol{p})= \mathop{min}\limits_{\boldsymbol{q}\in \mathcal{D}}||\boldsymbol{p}-\boldsymbol{{q}}||_2 - \mathop{min}\limits_{\boldsymbol{r}\in \mathcal{O}}||\boldsymbol{p}-\boldsymbol{r}||_2,
    \end{equation*}
    where $\mathcal{D}$ and $\mathcal{O}$ are the reconstruction results of \cite{dong2019multi} and ours.
    Since we pay more attention to the reconstruction of the objects, the reconstructed objects in our results have lower reconstruction error compared to \cite{dong2019multi}.
    Figure~\ref{fig:compare-to-dong-colormap} shows the error map on one representative scene. We can see that the method of \cite{dong2019multi} has larger reconstruction errors on objects with more geometric details.
}

On the other hand, we can see that our method is better according to all metrics, except for the distance consumption.
This is because that robots with \textit{\rev{reconstructor}} mode have to make some return trips in discovered areas around the incomplete objects to get better reconstruction results,
which leads to an increase in the total traveling distance consumption.
However, our method is still more efficient than the work of \cite{dong2019multi} even with longer traveling distance, reflected by our advantages in time consumption in the table.
Moreover, the time load balance is significantly lower than that of \cite{dong2019multi}.
The superiorities in time efficiency and load balances benefit from our asynchronous \textit{task-flow} model,
which makes sure that robots are occupied with tasks throughout the whole autoscanning process with no unemployed time.
Though the robots process tasks with different moving speeds, our distance load balance is similar to \cite{dong2019multi},
indicating that all robots process a similar workload in the whole scanning approach.

\begin{table}[t]
\centering
\caption{\rh{Comparison with the work of \cite{liu2018object} used in two different settings on reconstruction quality of objects and reconstruction efficiency.}}
\label{tab:compare-to-liu-efficiency}
\begin{tabular}{ccccccc}
\toprule
\multirow{2}{*}{Quality} & \multicolumn{3}{c}{Object Completeness} & \multicolumn{3}{c}{Object Accuracy}
\\ 
\cmidrule(r){2-4} \cmidrule(r){5-7}
& Small & Medium & Large & Small & Medium & Large
\\ 
\hline
NBO $\times$ 1 & 61.82 & \bf{72.91} & 69.76 & 0.033 & \bf{0.037} & 0.036
\\
NBO $\times$ 4 & \bf{67.52} & 71.71 & 69.92 & \bf{0.031} & 0.038 & 0.039
\\ 
 \bf{Ours} & 66.18 & 72.49 & \bf{70.03} & 0.035 & 0.039 & \bf{0.033}
\\
\bottomrule
\toprule
\multirow{2}{*}{Efficiency} & \multicolumn{3}{c}{Time Consumption} & \multicolumn{3}{c}{Distance Consumption}
\\ 
\cmidrule(r){2-4} \cmidrule(r){5-7}
& Small & Medium & Large & Small & Medium & Large
\\ 
\hline
NBO $\times$ 1 & 21.8 & 33.1 & 47.3 & 648.8 & 781.2 & 979.7
\\
NBO $\times$ 4 & 27.8 & 35.1 & 53.8 & 715.3 & 827.8 & 1128.7
\\ 
 \bf{Ours}& \bf{14.0} & \bf{18.9} & \bf{24.7} & \bf{536.1} & \bf{620.7} & \bf{848.5}
\\
\bottomrule
\end{tabular}
\end{table}

\begin{figure}[h]
\centering
\subfigure[NBO $\times$ 4]{
\includegraphics[width=0.47\linewidth]{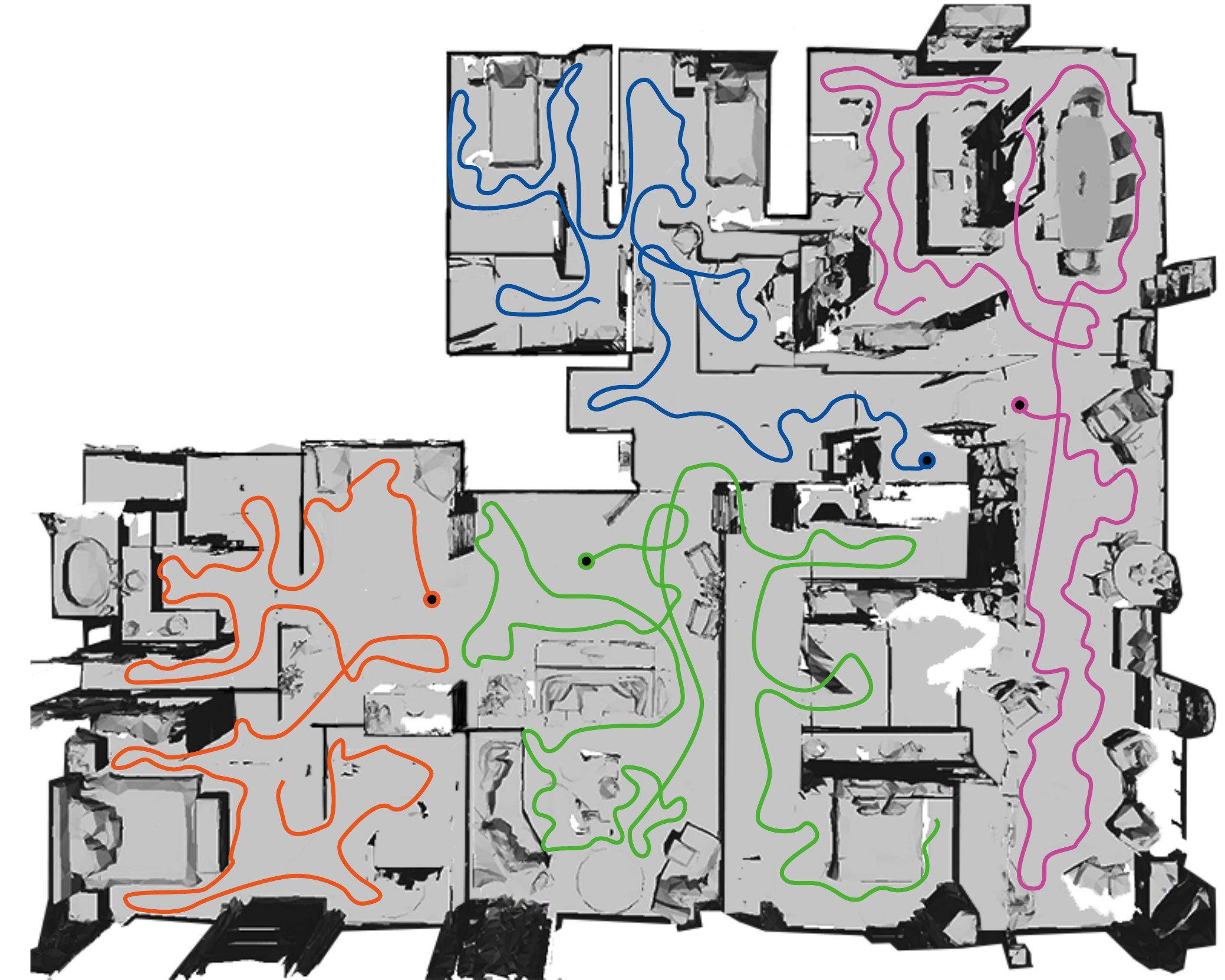}
}
\subfigure[Ours]{
\includegraphics[width=0.47\linewidth]{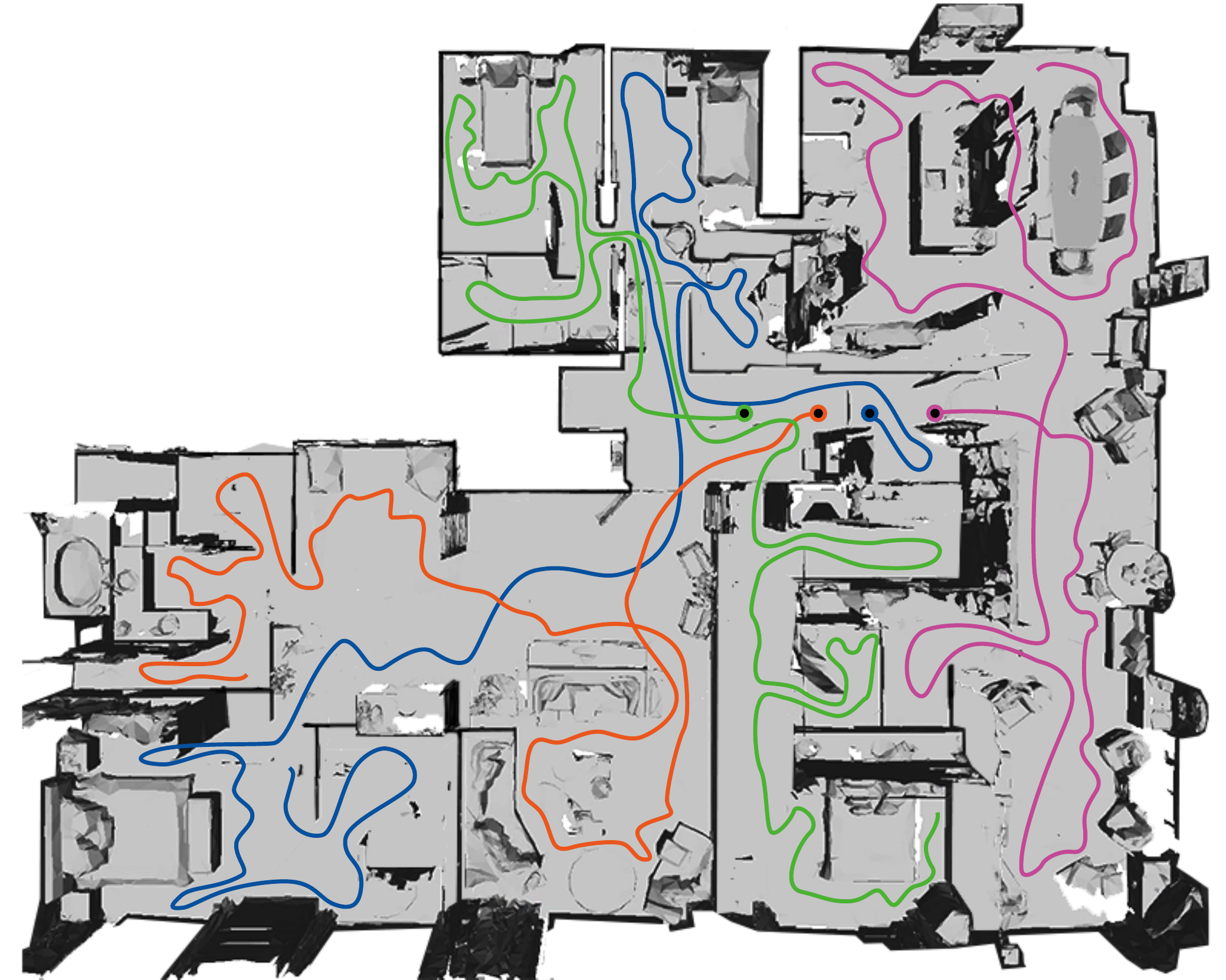}
}
\caption{\rh{Visual comparison between the result obtained using baseline NBO $\times$ 4 with multiple isolated robots and ours. 
The traveling paths of different robots are shown with lines in different colors.  }
}
\label{fig:compare-to-liu-path}
\end{figure}

When comparing those two baselines based on the work of \cite{liu2018object} with isolated robots,
our method gets better performance in both time and distance consumption when having comparable reconstruction quality, as shown in Table \ref{tab:compare-to-liu-efficiency}.
%
By using the same underlying method, the local robot paths are quite similar for baselines NBO $\times$ 1 and NBO $\times$ 4,
so we only show some visual comparison of our method to NBO $\times$ 4 in Figure \ref{fig:compare-to-liu-path}.
Note that multiple robots in our method can scan in various directions of the scene at a time, while for the NBO $\times$ 4 baseline,
each robot can only scan each pre-assigned sub-region without collaboration with others.
Moreover, since each robot only collects the local information of the scene, the robots spend a lot of time moving back and forth between different objects,
which leads to inefficient performance.

\ok{It is also an interesting result that NBO$\times$4 costs more time and distance than our baseline. The main reason is the underlying NBO method locates future tasks based on the objects glanced at during scanning, which leads to back-and-forth movement between different objects, while our method is able to jump out of local regions with the guidance of more global information collected from different robots. }

%


\subsection{Ablation studies}\label{ablation}

\paragraph{Ablation study on task assignment. } 
Compared to our dynamic task distribution, we freeze the robots' identities through all scanning periods.
We divide four robots into every possible combination,  denoted as as \textbf{NoSw(\rev{3E1R})}, \textbf{NoSw(\rev{2E2R})} and \textbf{NoSw(\rev{1E3R})}, to mainly check the change in the efficiency of the modified method.

The first three rows in Table \ref{tab:ablation-study} show the results of the experiments, and the result of our method is presented in the last row.
We can see that our full method beats all those variations with a salient advantage according to all the evaluation metrics.

With the decrease of the number of \textit{\rev{explorer}s}, the time consumption of the modified method rises, although the distance consumption keeps roughly the same, which is mainly due to the speed difference between the \textit{\rev{explorer}} and \textit{\rev{reconstructor}} modes.
With fewer \textit{\rev{explorers}} exploring the unknown region, few viewpoints for reconstruction can be generated in the limited discovered space and some of the \textit{\rev{reconstructors}} will keep waiting for new-discovered objects.


Moreover, the O-RMS decreases as the number of \textit{\rev{explorers}} decreases. Note that when there are fewer \textit{\rev{explorers}}, the completeness of the scanned objects is lower, so there are more reconstruction tasks generated for objects, which significantly increase the number of scans on objects and lead to lower O-RMS.

\begin{table}[t]

    \centering
    \caption{\rh{Ablation studies on different key components of our method.}
    }
    \label{tab:ablation-study}
    \begin{tabular}{ccccccc}
        \toprule
        Method           & T-C       & D-C        & O-Comp     & O-RMS       & D-LB       & T-LB
        \\
        \hline
        NoSw(\rev{3E1R}) & 19.6      & 698.8      & 58.91      & 0.0664      & 0.308      & 0.293
        \\
        NoSw(\rev{2E2R}) & 23.0      & 708.8      & 60.65      & 0.0438      & 0.932      & 0.326
        \\
        NoSw(\rev{1E3R}) & 27.6      & 676.7      & 59.89      & 0.0422      & 1.045      & 0.368
        \\
        \rev{NoRe}       & \bf{13.7} & 731.3      & 59.51      & 0.0538      & 0.194      & 0.085
        \\
        \rev{NoEx}       & 24.0      & 584.1      & 58.97      & 0.0494      & 0.170      & \bf{0.073}
        \\
        NoFlow           & 24.8      & \bf{536.1} & 58.54      & 0.0542      & 0.189      & 0.186
        \\
        \hline
        \bf{Ours}        & 19.3      & 668.4      & \bf{69.57} & \bf{0.0360} & \bf{0.169} & 0.079
        \\
        \bottomrule
    \end{tabular}
\end{table}

\begin{figure}[t]
    \centering
    \begin{overpic}[width=\linewidth,tics=5]{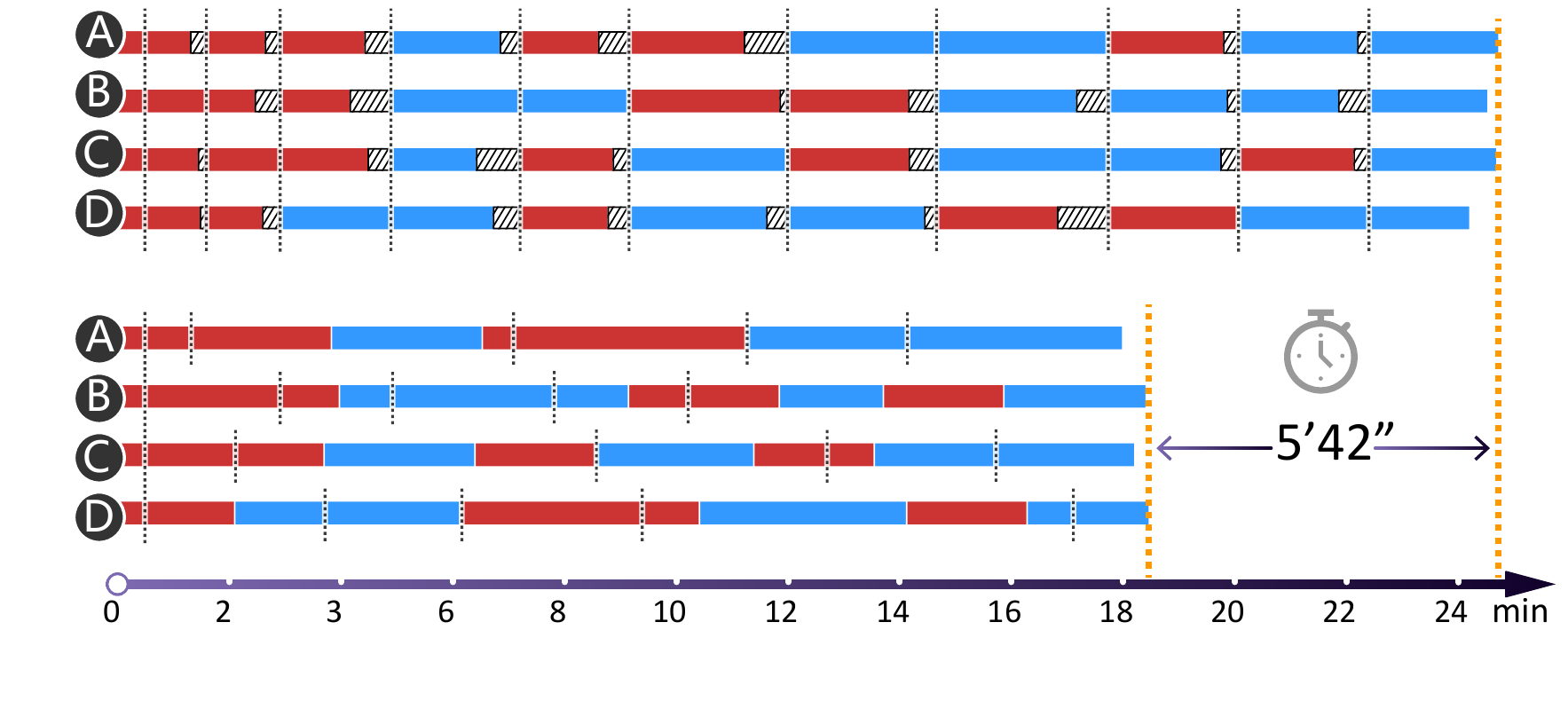}
        \put(1,31){\small \rotatebox{90}{NoFlow}}
        \put(1,14){\small \rotatebox{90}{Ours}}
        \put(37,0.7){\small Time Consumption}
    \end{overpic}
    \caption{
        \rh{ Comparison of the task scheduling between two settings of our method with (Ours) and without the asynchronous \textit{task-flow} model (NoFlow). 
            All robots switch between the \textit{\rev{explorer}} mode (red) and \textit{\rev{reconstructor}} mode (blue), and each time when the control center is activated for new task generation and assignment is indicated by the black dotted line.
            Comparing to our method, the robots in NoFlow have idle time (shown in black slash region) to wait for others finishing all assigned tasks, which leads to significant time waste during the whole process.
        }
    }
    \label{fig:compare-to-nosy-mode-switch}
\end{figure}





\paragraph{Ablation study on task execution.} 
We freeze the scanning mode of each robot to be either \textit{\rev{explorer}} or \textit{\rev{reconstructor}} no matter which type of tasks they are assigned, which results in two settings:
1) all \rev{explorer} properties (\textbf{\rev{NoRe}}), where all robots are equipped with the properties of high velocity and long scan range and only the camera on the head is utilized,
and 2) all \rev{reconstructor} properties (\textbf{\rev{NoEx}}), where all robots have the properties of low speed and short scan range using the camera in hand to scan.
The corresponding results are shown in the fourth and fifth rows in Table \ref{tab:ablation-study}.

With all \textit{\rev{reconstructor}} modes in \rev{NoEx}, the slow moving and scanning speed increase the overlapping of the scanning data, which significantly reduces the number of caves in the reconstruction result.
\ok{At the same time, it is worth pointing out that the O-RMS doesn't become lower when only using reconstructors.
    This is because of the accumulation of reconstruction errors when the reconstructors have to travel between objects during the scanning with their short and narrow visions. This error accumulation phenomenon has also been discovered in the previous work \cite{xu2017autonomous}. }
However, the distance consumption is lower than our method as our method will have more tasks during the scanning process with more incomplete objects scanned with \textit{\rev{explorer}} mode.
Nevertheless, due to the slow-moving speed, a significant amount of time is consumed during the scanning with all \textit{\rev{reconstructor}} modes compared to our method.

On the other hand, with all \textit{\rev{explorer}}  modes in \rev{NoRe}, the time consumption is much lower than our method due to the fast-moving speed of all robots, but this also leads to a considerable distance consumption.
To summarize, our method effectively combines these two scanning modes and gets balanced performance in scanning efficiency and reconstruction quality.

\begin{figure*}[htb]
  \centering
  \includegraphics[width=\linewidth]{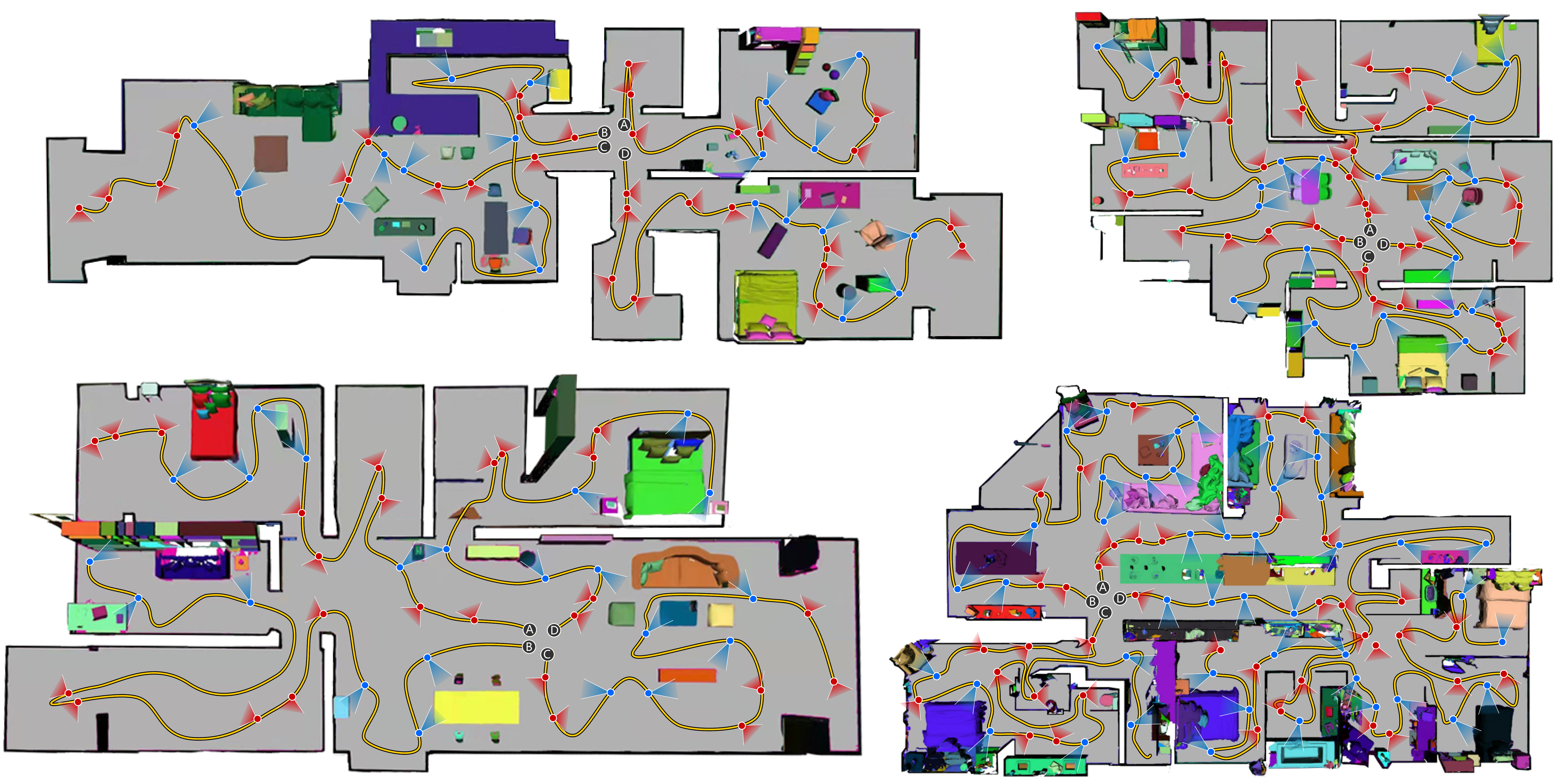}
  \caption{
    Some visual results of our method on four virtual scenes (three from the Front3D data and one from the MatterPort3D dataset).
    The traveling path is represented in yellow lines with black contours, with the exploration tasks (red triangles) and reconstruction tasks (blue triangles) shown on the top of the path. }

  \label{fig:final-result}
\end{figure*}

\paragraph{Ablation study on task scheduling.} 
    We remove the \textit{task-flow} model from our method, denoted as \textbf{NoFlow}, to test the effectiveness of the asynchronous task-scheduling module,
    where the calculation of the control center only triggers when all robots finish their task sequences assigned in the previous iteration.
    The comparison to our method can also be found in the last two rows in Table \ref{tab:ablation-study}.

    We can see that our method is better according to all metrics, except for distance consumption.
    This is because the new tasks generated overlap with some unfinished tasks due to the asynchronous module, which leads to more distance costs.
    But this small sacrifice of the distance gains the time-saving to a large level as no waiting is required during the scanning process,
    which results in generally more satisfactory results.

Figure \ref{fig:compare-to-nosy-mode-switch} shows the visual comparison of the task scheduling results of four robots when scanning the same scene.
    In our setting, more robots act as \rev{explorer}s in the very beginning when few objects are seen in the scene.
    After finishing several exploring tasks, some incomplete objects are found to trigger more reconstruction tasks.
    Besides, when almost all exploration tasks were finished, all robots changed their scanning modes into \rev{reconstructor}s to complete the last reconstruction tasks.
    For the method of NoFlow, the robots can only get new tasks after all robots finish their tasks; thus, there is a lot of waiting time, represented by the black slash interval in the figure.
    Moreover, this comparison also shows that the mode shift happens to every robot and each robot handles different tasks evenly in our method, which further proves the effectiveness of various scan modes.

\begin{figure*}[h]
	\centering
	\begin{overpic}[width=\linewidth,tics=5]{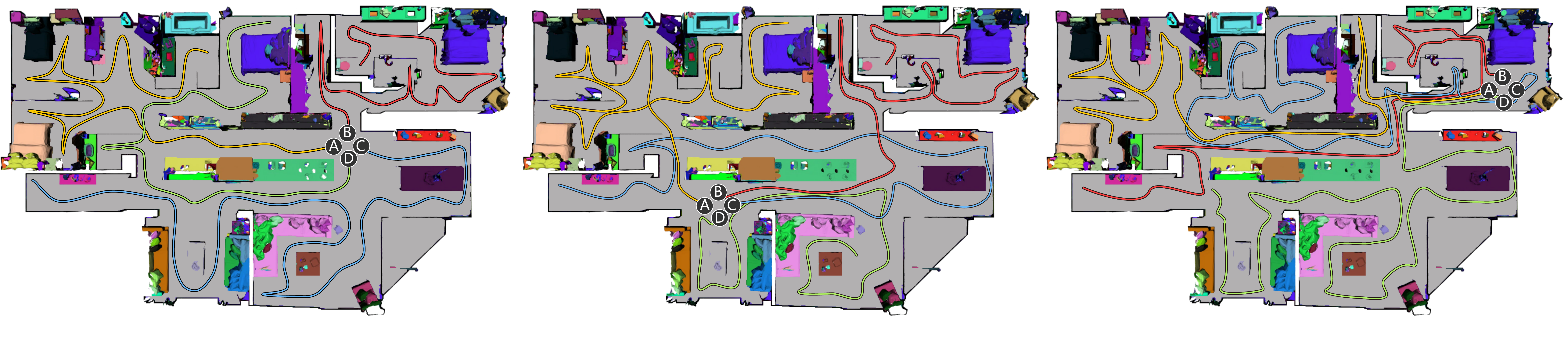}
		\put(16,0.4){\small (a)}
		\put(50,0.4){\small (b)}
		\put(84.5,0.4){\small (c)}
	\end{overpic}
\caption{
	\rh{Visualization of our results obtained with robots located at different initial points. 
	The starting points are depicted with black circles, and the traveling paths of different robots are shown in different colors.} }
\label{fig:different-init-locations}
\end{figure*}

\begin{figure}[htbp]
\centering
	\begin{overpic}[width=1.0\linewidth,tics=5]{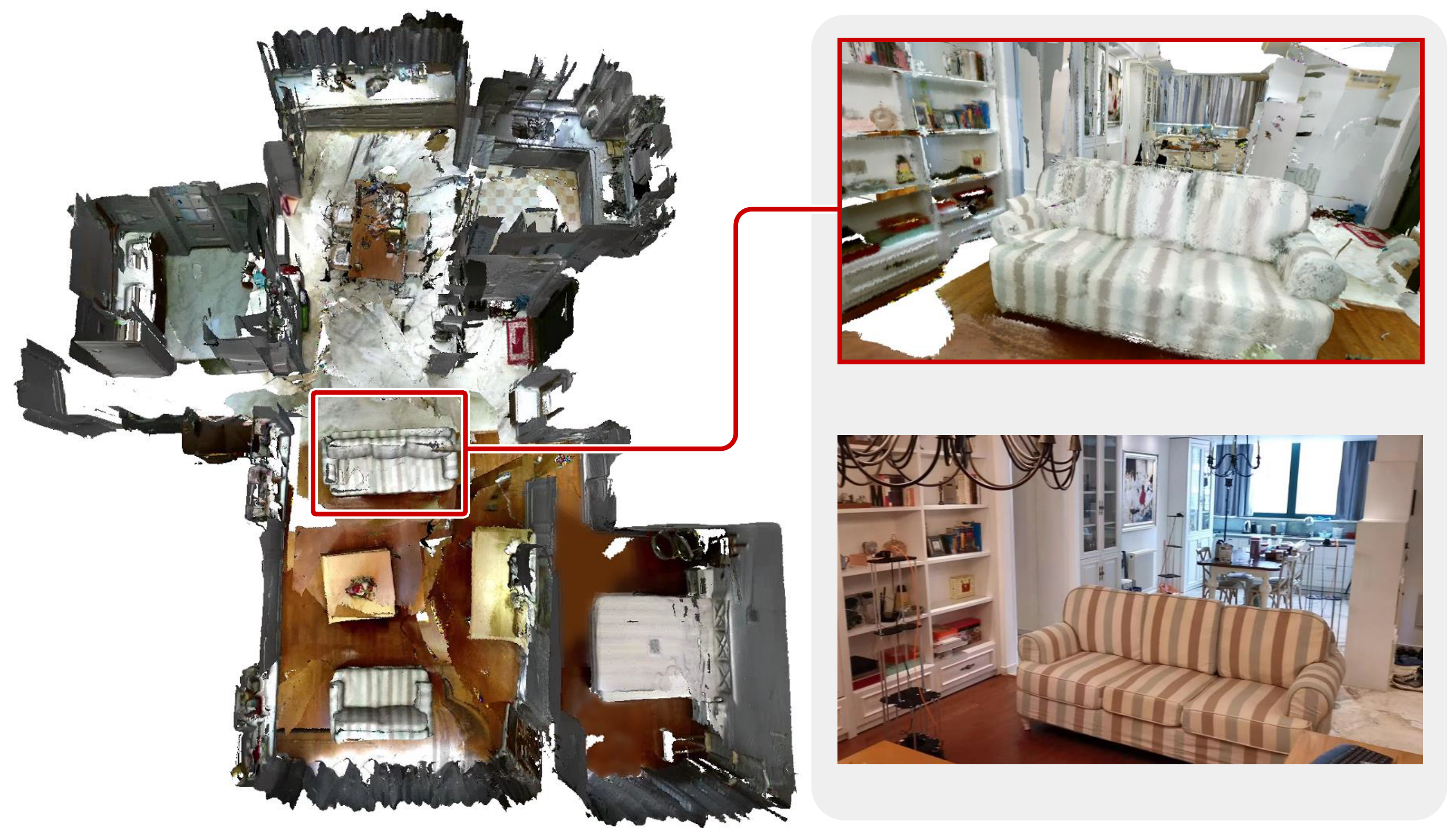}
		\put(68.5,29){\small Reconstruction}
		\put(72,1){\small Real Scene}
	\end{overpic}
	\caption{\rh{Example of the real-world scene reconstruction result with local details of the reconstruction result and corresponding picture taken from a similar viewpoint shown on the right.}} 
	\label{fig:reality-result}
\end{figure}

\subsection{Qualitative results}
\rh{\paragraph{Virtual simulations}
    Figure \ref{fig:final-result} shows the scanning paths and the semantic reconstruction results of our method for some exemplar indoor scenes.
    Different types of tasks are marked as viewpoints in different colors.
    We can see that our method can work well on indoor scenes with different scales and different structures.
    Different types of tasks are dynamically assigned to the robots for collaborative scanning during the whole process to maximize the exploration efficiency and reconstruction quality.}

\rh{Figure \ref{fig:different-init-locations}  further shows the results on the same scene but with different initial robot locations.
    We obverse that our method is quite robust to the initial locations, and the reconstruction quality and efficiency are comparable. }

\paragraph{Real robots tests}
\rh{
    We also test our method by scanning four unknown indoor scenes found around our department, including one rest area, one dorm, one area residence area, and one office.
    Figure \ref{fig:reality-result} shows one example reconstruction result, and other results can be found in the supplementary material.

    By using \texttt{Turtlebot3} in the real world, the scan result cannot achieve the same quality as the results in the virtual scenes due to the low degree-of-freedom for viewpoint controlling.
    There are also many inevitable complex environmental factors in the real world, such as lens jitters when moving and the position deviation.
    Although the final reconstruction quality is relatively lower than the results obtained in the virtual simulations, our method still shows its capability to reconstruct the scenes in the real world.
}

%

\section{Conclusions}
\rh{
    We present a multi-robot system for scene exploration, understanding, and object reconstruction tasks in unknown indoor scenes.
    To improve the exploration efficiency and the reconstruction quality of objects, we adopt the following optimization:
    The robots can change their scan modes dynamically according to the assigned tasks.
    Hence, the robots can possess suitable properties to accomplish their designated tasks efficiently with high quality.
    The robots with \textit{\rev{explorer}} mode explore the unknown areas of the scene, while the robots with \textit{\rev{reconstructor}} mode process the object reconstruction tasks.
    %
    %
    Moreover, we formulate the task assignment problem with multiple task varieties as a modified MDMTSP and approximate the optimal solution within a short time.
    %
    Furthermore, an asynchronous task-scheduling model is introduced to avert the periodic cycles in previous works to enhance the efficiency of the system.
    Finally, extensive experiments and comparisons are adopted to validate the feasibility and effectiveness of our algorithm.
}

\paragraph{Limitations.}
Our system for autoscanning suffers from several limitations.
First, there are times the proposed optimization method approaches the local optimal solution of the MDMTSP problem, which can decrease the efficiency of the final assignment results.
Second, our system can only be used in single-level indoor scenes, and the 2D occupancy grid cannot represent the complex scenes with compound structures.
Third, our system does not support the non-distributed solution.
When the scene is too large or complex for the control system to operate, robots may lose contact and stop receiving new tasks.
In that case, our system may face a significant failure with the robots out of control.
Besides, although we have prevented the robot from being scanned and reconstructed by others, the covering space of robots is not considered in the path planning stage.
There are also some failure situations when multiple robots are going through a narrow intersection together; the obstacle avoidance algorithm freezes them and leads to the system deadlock.

\paragraph{Future work.}
There are several directions to improve our method in the future.
First, task extraction can be updated with weights.
Not all viewpoints are equally important; for example, the viewpoints near a clean wall are less valuable than elaborate furniture.
Therefore, it is feasible to significantly improve reconstruction efficiency by gathering more task viewpoints around the complex surfaces.
Second, it will be an interesting future direction to generate the tasks using prior knowledge of the indoor scene's structure.
The layout of the indoor scene where the furniture is distributed in the different houses generally shares an implicit relationship.
Driving the robots towards places with furniture or more implicit unknown areas can improve scanning efficiency.
Finally, more types of sub-task can be introduced to our system.
For example, the object reconstruction task can be replaced by target object collection or target searching for other purposes.

\begin{acks}
   We thank the anonymous reviewers for their valuable comments.
   This work was supported in parts by
   the National Natural Science Foundation of China (61872250, 62025207),
   Guangdong Natural Science Foundation (2021B1515020085),
   Shenzhen Science and Technology Program (RCYX20210609103121030),
   and Guangdong Laboratory of Artificial Intelligence and Digital Economy (SZ).
\end{acks}

\bibliographystyle{ACM-Reference-Format}
\bibliography{AsyncScan}

\end{document}